\documentclass[10pt,twocolumn,letterpaper]{article}

\usepackage{cvpr}              
\usepackage{amsmath}
\usepackage{multirow}
\usepackage{booktabs}
\usepackage[table]{xcolor}
\usepackage{xcolor}
\usepackage{makecell}
\usepackage{pifont}
\usepackage{textcomp} 
\definecolor{cvprblue}{rgb}{0.21,0.49,0.74}
\definecolor{mygray}{HTML}{E0E0E0}
\definecolor{mygreen}{HTML}{C8E6C9}
\definecolor{myorange}{HTML}{FFE0B2}
\usepackage[pagebackref,breaklinks,colorlinks,allcolors=cvprblue]{hyperref}

\def\paperID{*****} 
\def\confName{CVPR}
\def\confYear{2026}

\begin{document}
\title{SparseOccVLA: Bridging Occupancy and Vision-Language Models via Sparse Queries for Unified 4D Scene Understanding and Planning}


\author{
Chenxu Dang\textsuperscript{1,2,3}\thanks{Work done during the internship at Xiaomi EV and AIR.} \quad
Jie Wang\textsuperscript{2} \quad
Guang Li\textsuperscript{2} \quad
Zhiwen Hou\textsuperscript{2} \quad
Zihan You\textsuperscript{3} \quad
Hangjun Ye\textsuperscript{2} \\
Jie Ma\textsuperscript{1} \quad 
Long Chen\textsuperscript{\rm 2}\thanks{Corresponding Authors.} \quad
Yan Wang\textsuperscript{\rm 3}\footnotemark[\value{footnote}]\\[1mm]
\textsuperscript{1}Huazhong University of Science and Technology \quad \\
\textsuperscript{2}Xiaomi EV \quad 
\textsuperscript{3}Institute for AI Industry Research (AIR), Tsinghua University\\
{ \url{https://msundyy.github.io/SparseOccVLA}}
}

\maketitle


\begin{abstract}
In autonomous driving, Vision Language Models (VLMs) excel at high-level reasoning , whereas semantic occupancy provides fine-grained details. Despite significant progress in individual fields, there is still no method that can effectively integrate both paradigms. Conventional VLMs struggle with token explosion and limited spatiotemporal reasoning, while semantic occupancy provides a unified, explicit spatial representation but is too dense to integrate efficiently with VLMs. To address these challenges and bridge the gap between VLMs and occupancy, we propose \textbf{SparseOccVLA}, a novel vision-language-action model that unifies scene understanding, occupancy forecasting, and trajectory planning powered by \textbf{sparse occupancy queries}. Starting with a lightweight Sparse Occupancy Encoder, SparseOccVLA generates compact yet highly informative sparse occupancy queries that serve as the single bridge between vision and language. These queries are aligned into the language space and reasoned by the LLM for unified scene understanding and future occupancy forecasting. Furthermore, we introduce an LLM-guided Anchor-Diffusion Planner featuring decoupled anchor scoring and denoising, as well as cross-model trajectory-condition fusion. SparseOccVLA achieves a 7\% relative improvement in CIDEr over the state-of-the-art on OmniDrive-nuScenes, a 0.5 increase in mIoU score on Occ3D-nuScenes, and sets state-of-the-art open-loop planning metric on nuScenes benchmark, demonstrating its strong holistic capability.
\end{abstract}    
\section{Introduction}
\label{sec:intro}

In recent years, Vision-Language Models (VLMs)\cite{qwen-vl,llava,internvl,blip,blip2,clip} have witnessed remarkable progresses in end-to-end autonomous driving\cite{uniad,vad,sun2025sparsedrive,drivetransformer}. With strong understanding and reasoning capabilities, VLMs are expected to handle long-tail cases and offer interpretability.

However, conventional VLMs pretrained on web-scale 2D images inherently lack temporal-spatial awareness\cite{mpdrive,surds}. Autonomous driving systems are required to process multi-view, multi-frame video streams for comprehensive perception, which imposes an astronomical number of vision tokens and impedes the deployment of VLMs. Some approaches\cite{omnidrive,solve} adopt Q-Former\cite{blip2} to compress tokens, but their indiscriminate global cross-attention inevitably discards image details, resulting in suboptimal performance. Moreover, multi-view cameras introduce view ambiguity\cite{senna}, which remains a significant challenge.

Recent studies\cite{hermes,opendrivevla} project multi-view images into Bird’s-Eye View (BEV). Although alleviating view ambiguity, the dense BEV still incurs an enormous number of tokens (e.g., a $50\times50$ BEV map corresponds to 2,500 tokens). In practice, due to inherent sparsity of road scenes, more than 80\% of BEV tokens are invalid, revealing the inefficiency of BEV representation. Moreover, the compression of BEV inevitably sacrifices spatial details, making it difficult to handle unstructured scenarios.

Compared to BEV, semantic occupancy offers more comprehensive and granular representations\cite{occnet}, and has attracted increasing attention\cite{cotr,sparseocc,panoocc,flashocc} since its introduction. Nevertheless, occupancy and VLMs have long evolved separately, with no effective integration achieved to fully exploit the strengths of both. The key challenge lies in the inherently dense and low-level nature of occupancy, which hinders its alignment with LLMs. Some early attempts\cite{occllama,occllm} discretized occupancy ground truth and embedded it via VQ-VAE\cite{vqvae}, but such paradigms separate understanding from perception and discards critical non-geometric visual cues—such as traffic lights, lane markings and signs. Recent OccVLA\cite{occvla} employs occupancy supervision to enhance VLMs, yet still relies on visual tokens and thus inherits the limitations of conventional VLMs. Overall, occupancy remains marginal within VLMs, and how to integrate them effectively has long perplexed the community.

Recently, sparse representations have achieved great success in occupancy perception\cite{opus,sparseocc} and forecasting\cite{sparseworld}, providing a promising avenue for integrating occupancy with VLMs. In response, we propose SparseOccVLA. Relying solely on sparse occupancy queries, SparseOccVLA effectively bridging low-level occupancy perception with high-level scene understanding.

SparseOccVLA starts with a Sparse Occupancy Encoder, where learnable query embeddings progressively interact with multi-view, multi-scale image features through stacked encoding layers, ultimately generating compact occupancy queries (only a few hundred). These queries are aligned with the language space, serving as the sole bridge between visual input and the LLM. We also specifically design token-level distillation and global query mechanism to facilitate the cross-modal alignment. The resulting multi-modal tokens reasoned by a unified LLM are utilized in parallel for scene understanding and future occupancy forecasting.

To investigate the capability of our sparse occupancy–oriented VLM in downstream planning, we design an LLM-guided Anchor-Diffusion planner where the LLM is prompted to assign high-level scores to trajectory anchors. A diffusion decoder then allows text-level instruction features, occupancy queries and ego states to interact alternately with noised trajectories, enabling cross-modal trajectory-condition fusion and noisy prediction. This decoupling exploits the complementary strengths of LLMs for decision-making and diffusion models for regression.

Extensive experiments demonstrate the strong holistic capability of SparseOccVLA. Notably, it attains a CIDEr score of 0.795 on the OmniDrive-nuScenes\cite{omnidrive} benchmark, surpassing the recent HERMES\cite{hermes} by 7\%. For occupancy forecasting, our method outperforms the recent SparseWorld\cite{sparseworld} by up to 0.51 in average mIoU. Moreover, our planner establishes new SOTA performance on open-loop metrics in the nuScenes dataset. 

We summarize our key contributions as follows:
\begin{itemize}
    \item To the best of our knowledge, SparseOccVLA is the first bonafide occupancy-oriented end-to-end VLA model that unifies understanding, forecasting and planning.
    \item The compact and information-rich sparse queries effectively link occupancy representation with scene understanding, serving as a novel vision-language connector superior to conventional ones for autonomous driving.
    \item We design an LLM-guided anchor diffusion planner, which simultaneously leverages the strengths of both LLMs and diffusion models.
    \item SparseOccVLA achieves strong performance in scene understanding, occupancy prediction, and planning tasks.
\end{itemize}


    


\section{Related Work}
\subsection{VLM for Autonomous Driving}
Early end-to-end planners\cite{uniad,vad,sun2025sparsedrive,sparsead} are generally regarded as black-boxes, lacking interpretability and struggling to handle long-tail scenarios. In recent years, vision-language models\cite{blip,qwen-vl,llava} (VLMs) have been widely adopted for scene understanding\cite{drivelm,drivelmm,omnidrive} and chain-of-thought (CoT)\cite{RAD-Driver,reason2drive,x-driver} reasoning. Subsequent works sought to integrate VLMs with end-to-end planners into dual-system architectures\cite{senna,solve,drivevlm}.

To align images with LLMs, some methods\cite{senna,x-driver,drivemonkey} employ a standard MLP projection, which can produce excessively large and cumbersome visual tokens. Others\cite{omnidrive,solve,orion,reason2drive} employ a Q-Former\cite{blip2} mechanism, which utilizes a small number of queries to attend to the entire image for token compression. Such indiscriminate global interaction, however, inevitably results to the loss of rich semantic details. To enhance the spatial understanding capability of VLMs, recent approaches\cite{hermes,opendrivevla} attempt to project multi-view images into the BEV space before feeding them into LLMs. However, BEV representations suffer from uneven information distribution.

\subsection{Occupancy for Autonomous Driving}
Semantic occupancy provides a more detailed and fine-grained spatial representation\cite{occnet}, particularly excelling at modeling irregular elements. Early studies mainly focused on occupancy perception\cite{flashocc,cotr,panoocc,occformer} and static forecasting\cite{cam4docc,efficientocf}. OccWorld\cite{occworld} and its successors\cite{dome,preworld,i2world} introduced occupancy-based world models to autoregressively forecast future occupancies and trajectories. To address the intrinsic density of occupancy, several sparse approaches\cite{opus,sparseocc,sparseworld} have been proposed, enhancing their applicability in downstream tasks.

\subsection{Combination of occupancy and VLMs}
In the era of large language model, some works\cite{occllama,occllm} attempted to discretize occupancy ground truth and embed via VQ-VAE\cite{vqvae} for LLM understanding. However, such designs isolate the LLM from the perception, completely discarding non-geometric visual cues—such as traffic lights and lane markings—that are critical for driving decisions. The concurrent OccVLA\cite{occvla}, leverages occupancy supervision to enhance VLA, yet it still relies on visual tokens.

Despite the remarkable success of both occupancy and VLMs, to the best of our knowledge, no existing work has effectively integrated them. This is primarily due to occupancy’s dense nature creating a significant modality gap. A more detailed analysis is included in the supplementary material, which we encourage readers to consult.

\begin{figure*}
    \centering
    \includegraphics[width=1\linewidth]{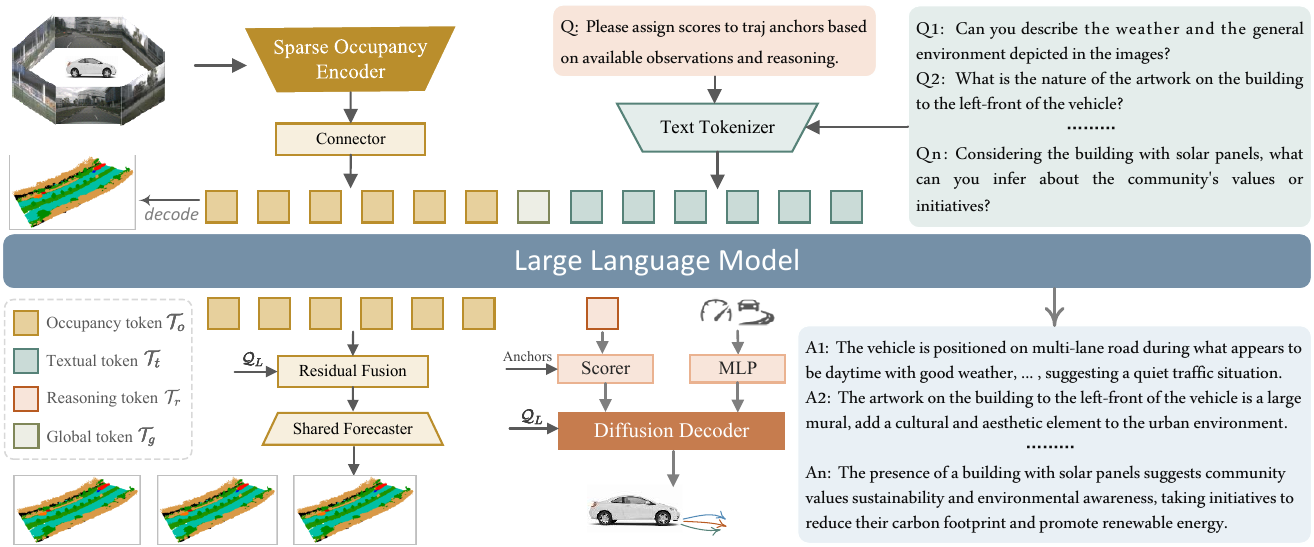}
    \caption{The overview of SparseOccVLA, which is based on sparse occupancy queries, with a large language model serving as a unified token processor to simultaneously perform scene understanding, occupancy forecasting, and trajectory planning.}
    \label{fig:overview}
\end{figure*}
\section{Method}
The overall architecture of our SparseOccVLA is illustrated in Fig. \ref{fig:overview}. The inputs include multi-view, multi-frame images, textual instructions, and ego-state information. The images are processed by a Sparse Occupancy Encoder (Sec. \ref{sec:sparse occupancy encoder}), which generates a set of sparse occupancy queries. With no visual tokens, only aligned occupancy tokens are fed into the LLM (Sec. \ref{sec:unified large language model}) for unified understanding and forecasting. In(Sec. \ref{sec:anchor-based diffusion planner}), we elaborate on the LLM-guided Anchor-Diffusion Planner for cross-modal fusion with trajectory anchors.

\subsection{Sparse Occupancy Encoder}
\label{sec:sparse occupancy encoder}
\begin{figure}
    \centering
    \includegraphics[width=1.0\linewidth]{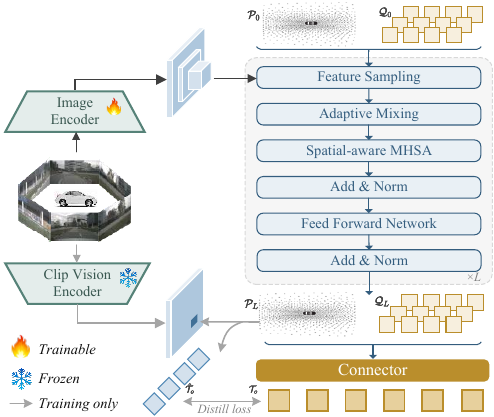}
    \caption{Illustration of the Sparse Occupancy Encoder, where the distillation branch (indicated by \textcolor{gray}{gray}) is removed during inference.}
    \label{fig:occupancy_encoder}
\end{figure}

As shown in Fig. \ref{fig:occupancy_encoder}, Sparse Occupancy Encoder takes as input multi-view, multi-frame images $\mathcal{I}=\{I_{v,f}|v=1,...,V;f=1,...,F\}$. We randomly initialize a set of sparse occupancy query embeddings $\mathcal{Q}_0=\{{\mathbf{q}_i\in \mathbb{R}^D}\}_{i=1}^N$, along with their 3D positions $\mathcal{P}_0 = \{\mathbf{p}_i\in\mathbb{R}^3\}_{i=1}^N$. Here, $F$, $V$, $N$, and $D$ represent the number of image frames, camera views, queries, and feature dimension, respectively.

We employ a lightweight image encoder to extract multi-scale features for each image. The initial occupancy embeddings $[\mathcal{Q}_0,\mathcal{P}_0]$ are fed into a stack of $L$ encoding layers. The encoding layer follows the design of SparseBEV\cite{sparsebev}, consisting of feature sampling, adaptive fusion, spatial-aware multi-head self-attention (Spatial-aware MHSA), and a FFN. This design effectively extracts cross-frame and multi-scale features and has been widely adopted in recent works\cite{sparseocc,opus,sparseworld}. Following each layer, the queries are decoded into a set of occupancy points while updating their coordinates consequently. We progressively increase the number of points predicted across layers, enabling a coarse-to-fine occupancy encoding. The point set $\mathcal{P}_l$ output from each layer is supervised with Chamfer Distance (CD):

\begin{equation}
    \text{CD}(\mathcal{P}_l,\mathcal{G}) = \sum_{\textbf{p} \in \mathcal{P}_l} \min_{\textbf{g} \in \mathcal{G}} \|\textbf{p} - \textbf{g}\|_2^2 + \sum_{\textbf{g} \in \mathcal{G}} \min_{\textbf{p} \in \mathcal{P}_l} \|\textbf{p} - \textbf{g}\|_2^2,
\label{eq: chamfer}
\end{equation}
where $l=1,2,...,L$ and $\mathcal{G}$ denotes the sparsified ground-truth point set. The semantic output of each predicted point is supervised using the focal loss with respect to the label of its matched ground-truth point.

The outputs $\mathcal{Q}_L$ and $\mathcal{P}_L$ from the final $L$-th encoding layer are combined and aligned to the LLM space through a lightweight connector (implemented as an MLP):
\begin{equation}
\begin{aligned}
    \mathcal{T}_o&=\text{MLP}(\mathcal{Q}_L+\text{PE}(\mathcal{P}_L)),
\end{aligned}
\end{equation}
where, PE denotes positional encoding. The aligned representations $\mathcal{T}_o$, referred to as occupancy tokens, inherently encode geometric and semantic cues in a highly compact form, allowing them to serve as a complete substitute for conventional image tokens for LLM understanding. 

\paragraph{Feature-level Distillation}

In our early exploration, we observed that the LLM exhibited an extremely high initial loss, which led to slow convergence or even training collapse. We attribute this to the large gap between the low-level occupancy representation and the high-level language space. This is particularly exacerbated by the sparsity and disorder of the occupancy queries.

To address this issue, we design a feature-level distillation loss to assist training. As in Fig. \ref{fig:occupancy_encoder}, when training the occupancy encoder, we simultaneously extract single-frame, single-scale feature from a pretrained and frozen CLIP\cite{clip} Vision Encoder. The occupancy tokens $\mathcal{T}_o$ are projected via their 3D coordinates and to sample the teacher features $\hat{\mathcal{T}_o}$ via interpolation. A cosine similarity loss is computed between $\mathcal{T}_o$ and $\hat{\mathcal{T}_o}$ to encourage alignment.

However, we observe that overly strict constraints significantly hinders the occupancy encoder learning. Therefore, before computing the loss, we apply independent LayerNorm layers to $\mathcal{T}_o$ and $\hat{\mathcal{T}_o}$, which normalize their mean and variance and align only their relative feature shapes. Moreover, the learnable $\gamma$ and $\beta$ further mitigate the strength of the constraint. Consequently, the final distillation loss is formulated as:
\begin{equation}
    \mathcal{L}_{\text{distill}} = 1-\text{cosine}(\text{Norm1}(\mathcal{T}_o),\text{Norm2}(\hat{\mathcal{T}}_o)).
\end{equation}

\subsection{Unified Large Language Model}
\label{sec:unified large language model}
\paragraph{Inputs}
As discussed in Sec. \ref{sec:sparse occupancy encoder}, the occupancy tokens preserve rich details but focus only on local regions, lacking a global representation. To mitigate this limitation, we introduce a small set of learnable global scene tokens $\mathcal{T}_g$ that query all occupancy tokens $\mathcal{T}_o$ via cross-attention:
\begin{equation}
    \mathcal{T}_g=\text{Cross\_Attn}(\mathcal{T}_g,\mathcal{T}_o,\mathcal{T}_o).
\end{equation}

The language instructions are tokenized as textual tokens $\mathcal{T}_t$. Ultimately, the combined tokens $\mathcal{T}_{all}=[\mathcal{T}_o,\mathcal{T}_g, \mathcal{T}_t]$ are sequentially fed info the LLM for processing and joint execution of downstream understanding and forecasting.

\paragraph{Understanding}
Following prior works\cite{omnidrive,hermes}, our SparseOccVLA autoregressively generates answers based on perception-level tokens and textual prompts, describing the driving environment, localizing key objects, and making high-level driving decisions. The LLM is supervised using the standard maximum likelihood estimation (MLE) loss:
\begin{equation}
\mathcal{L}_{\text{LM}} = - \sum_{n=1}^{N} \log P_\theta \big( y_t \mid y_{<n} \big),
\end{equation}
where $N$ denotes the token length.

For simplicity, we employ causal attention over all tokens. It is noteworthy that, unlike visual tokens that are structured, occupancy tokens are essentially point clouds, which are sparse and unordered, thus the LLM must rely solely on their 3D coordinates to infer spatial topology. We observe that manually permuting the order of occupancy tokens or using bidirectional masking does not promote overall performance, which is counterintuitive, but also demonstrates that the LLM can understand the scene relying solely on token positions, enhancing overall robustness.

\paragraph{Forecasting}

Inspired by \cite{hermes}, we consider that  occupancy tokens $\mathcal{T}_o$ after LLM reasoning, denoted as $\mathcal{T}_o'$, carry linguistic cues and a global context, while the occupancy queries $\mathcal{Q}_L$ retain low-level details, their combination may enable complementary interaction. Consequently, we design a simple residual fusion linear layer to merge them:
\begin{equation}
    \hat{Q}_o=\text{MLP}([\mathcal{T}_o',\mathcal{Q}_L]). 
\end{equation}

Here, $\hat{Q}_0$ and $\mathcal{Q}_L$ share the same dimension $D$. Furthermore, $\hat{Q}_o$ are progressively augmented with 4D embeddings and ego-vehicle states:
\begin{equation}
    \hat{Q}^t_o=\hat{Q}_o+\text{embed}(\text{X,Y, Z},t)+\text{Ego},
\end{equation}
where X, Y, Z denote the 3D coordinates of queries, and $t=1,2,...,T$, $T$ represents the horizon to be forecasted.
A shared forecaster is employed to recursively forecast the future occupancy frame by frame, which is supervised with the corresponding frame's occupancy ground truth, consistent with Sec. \ref{sec:sparse occupancy encoder}.

\subsection{LLM-guided Anchor-Diffusion Planner}
\label{sec:anchor-based diffusion planner}

Recently, anchor-based diffusion planner\cite{diffusiondrive} has demonstrated significant advantages, where diffusion is applied to slightly perturb anchors, followed by per-anchor scoring and denoising. However, these methods still lack integration with LLM reasoning and guidance.

To bridge this gap, we propose an LLM-guided anchor-diffusion planner. Following \cite{diffusiondrive}, we first apply K-means clustering on the training trajectories to obtain $K$ anchor trajectories $\{\mathbf{a}_k\}_{k=1}^K$, which naturally exhibit significant diversity. As in Fig. \ref{fig:overview}, We prompt the LLM to generate reasoning tokens $\mathcal{T}_r$, followed by a scorer that assigns scores to anchors based on contextual understanding, reasoning, and ego state. The scorer is supervised with BCE Loss, where the anchor closest to the ground-truth is regarded as the positive sample and the others as negatives.

We follow the standard DDIM\cite{ddim} noising and denoising strategy to optimize the anchors. During training, Gaussian noise is add to all anchors:
\begin{equation}
\tau_k^i = \sqrt{\bar{\alpha}^i}\mathbf{a}_k + \sqrt{1 - \bar{\alpha}^i}\,\boldsymbol{\epsilon}, 
\quad \boldsymbol{\epsilon} \sim \mathcal{N}(0, \mathbf{I}),
\end{equation}
where $i$ denotes the noising step. The noised anchors $\{\tau_k\}_{k=1}^K$ are jointly fed into a conditional diffusion decoder to regress the clean trajectories directly. Only the positive anchor is supervised using $L1$ loss. During inference, the anchor trajectories are randomly noised and iteratively denoised, and the denoised result corresponding to the anchor selected by the LLM is chosen as the final trajectory.
\begin{figure}
    \centering
    \includegraphics[width=1.0\linewidth]{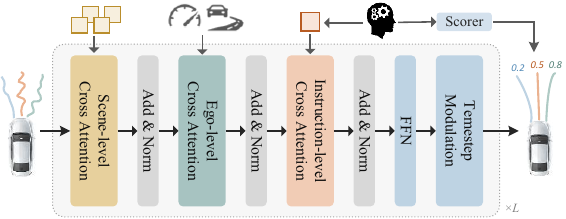}
    \caption{The architecture of Diffusion Decoder.}
    \label{fig:planner}
\end{figure}

As shown in Fig. \ref{fig:planner}, the diffusion decoder encodes the noised trajectories and employ stacked cross-attention layers to sequentially attend to (1) scene-level occupancy queries $\mathcal{Q}_0$, (2) the encoded ego state, and (3) the LLM-generated tokens for decision. This multi-modal and cross-level fusion enables the planner to jointly reason over geometric perception, dynamic context, and high-level semantics, thereby improving the rationality and consistency of the generated trajectories. Here, we completely discard BEV features, boxes and maps, only occupancy queries serving as perceptual input. A feed-forward network (FFN) and a timestep modulation layer are applied to enhance temporal conditioning and representation stability.

\section{Experiments}
\subsection{Datasets and Benchmarks}
We benchmark our model on the public nuScenes dataset\cite{nuscenes}, which contains 700/150/150 scenes for training, validation, and testing. Each scene provides six camera views and spans 20 seconds with 2 Hz key frames.

For understanding, we adopt OmniDrive-nuScenes\cite{nuscenes} benchmark, which provides detailed captions and VQA pairs generated by GPT-4V. Following prior works\cite{omnidrive,hermes}, we report METEOR\cite{meteor}, CIDEr\cite{cider}, and ROUGE\cite{rouge} metrics. For occupancy forecasting task, we employ Occ3D-nuScenes\cite{occ3d}, which divides the 3D space into a $200\times200\times16$ voxel grid with a 0.4m resolution. It provides high-quality dense semantic annotations covering 18 categories (including free class). We evaluate SparseOccVLA's forecasting performance through the mean Intersection over Union (mIoU) over the future 3 seconds. To ensure a fair comparison, we do not utilize camera mask while computing the mIoU. For the planning task, the evaluation metrics include open-loop L2 error and collision rate over future 3s. We compute understanding and planning scores over the full dataset. For the mIoU forecasting, the first five and last seven frames of each scene are excluded, consistent with prior works\cite{occworld,preworld}.

\begin{table*}[htbp]
\centering
\setlength{\tabcolsep}{8pt} 
\renewcommand{\arraystretch}{1.2} 
\caption{Performance comparison of SparseOccVLA with current state-of-the-art methods for scene understanding (left) and occupancy prediction (right). The best results are highlighted in \textbf{bold}. "Token" refers to the number of tokens representing image information.}
\label{tab:understanding and forecasting}
\small
\begin{tabular}{llccc c@{\hspace{0.25mm}} c@{\hspace{0.25mm}} lcccc}
\toprule
\multicolumn{5}{c}{\textbf{Left Table: Understanding}} & &&  \multicolumn{5}{c}{\textbf{Right Table: Forecasting}} \\
\midrule
\rule{0pt}{0.4cm}  \multirow{2}{*}{Method} & \multirow{2}{*}{Token} & \multicolumn{3}{c}{Understanding$\ \uparrow$}&& &\multirow{2}{*}{Method} &  \multicolumn{4}{c}{mIoU$\ \uparrow$}  \\
\cmidrule(rl){3-5} \cmidrule(rl){9-12}
 &  & METEOR & ROUGE & CIDEr & && & 1s & 2s & 3s & Avg  \\
\midrule
LLaVA-OV\cite{llava} & $>$1500       & -     & 0.221 & 0.284& && OccWorld\cite{occworld} &  11.55 & 8.10 & 6.22 & 8.62 \\
OmniDrive-BEV\cite{omnidrive} & 2500 & 0.356 & 0.278 & 0.595&& &OccLLaMA\cite{occllama} &  10.34 & 8.66 & 6.98 & 8.66 \\
OmniDrive\cite{omnidrive} & 512 & 0.380 & 0.326 & 0.686 & & &PreWorld\cite{preworld}& 12.27 & 9.24 & 7.15 & 9.55 \\
HERMES-p\cite{hermes} & 2500 & 0.380 & 0.323 & 0.726& & &Occ-LLM\cite{occllm} &   11.28 & 10.21 & 9.13 & 10.21 \\
HERMES\cite{hermes} & 2500 & 0.384 & 0.327 & 0.741& & &SparseWorld\cite{sparseworld} &  14.93 & 13.15&  11.51 & 13.20  \\
\midrule

\cellcolor{myorange}SparseOccVLA & \cellcolor{myorange}300 & \cellcolor{myorange}0.384 &\cellcolor{myorange} 0.333 & \cellcolor{myorange}0.762 & & & \cellcolor{mygreen}SparseOccVLA & \cellcolor{mygreen}15.36 & \cellcolor{mygreen}13.12 & \cellcolor{mygreen}11.42 & \cellcolor{mygreen}13.30 \\
\cellcolor{myorange}SparseOccVLA & \cellcolor{myorange}600 & \cellcolor{myorange}\textbf{0.388} &\cellcolor{myorange} \textbf{0.337} & \cellcolor{myorange}\textbf{0.796} & & & \cellcolor{mygreen}SparseOccVLA& \cellcolor{mygreen}\textbf{15.86} & \cellcolor{mygreen}\textbf{13.54} & \cellcolor{mygreen}\textbf{11.72} & \cellcolor{mygreen}\textbf{13.71} \\
\bottomrule
\end{tabular}
\end{table*}

\begin{table*}[htbp]
\centering
\caption{Comparison of SparseOccVLA with different types of planning models on open-loop planning performance in nuScenes. The best results are highlighted in \textbf{bold}.}
\label{tab:planning}
\setlength{\tabcolsep}{5pt} 
\renewcommand{\arraystretch}{1.00} 

\resizebox{\textwidth}{!}{ 
\small
\begin{tabular}{l c c cccc cccc}
\toprule
\multirow{2}{*}{\textbf{Method}} & \multirow{2}{*}{\textbf{Type}} & \multirow{2}{*}{\textbf{Auxiliary Supervision}} &
\multicolumn{4}{c}{\textbf{L2 (m)$\downarrow$}} & \multicolumn{4}{c}{\textbf{Collision Rate (\%)$\downarrow$}} \\
\cmidrule(lr){4-7} \cmidrule(lr){8-11}
 & & & 1s & 2s & 3s & Avg. & 1s & 2s & 3s & Avg. \\
\midrule
UniAD\cite{uniad} & End to End &Map\ \&\ Box\ \&\ Motion\ \&\ Occ &0.20 &0.42 &0.75 &0.46 &0.02 &0.25 &0.84 &0.37 \\
VAD-Base\cite{vad} &End to End &Map\ \&\ Box\ \&\ Motion &0.17 &0.24 &0.60 &0.37 &0.04 &0.27 &0.67 &0.33 \\
\midrule
OccWorld\cite{occworld} & World model &3D Occ &0.39 &0.73& 1.18 &0.77 &0.11 &0.19 &0.67 &0.32 \\
PreWorld~\cite{preworld} & World model & 2D Label\ \&\ 3D Occ & 0.22 & 0.30 & 0.40 & 0.31 & 0.21 & 0.66 & 0.71 & 0.53 \\
SparseWorld~\cite{sparseworld} & World model & 3D-Occ & 0.19 & 0.25 & 0.36 & 0.27 & 0.11 & 0.29 & 0.46 & 0.29 \\

\midrule
HERMES-p~\cite{hermes} & VLM & QA\ \&\ Lidar & 0.16 & 0.32 & 0.59 & 0.36 & 0.00 & 0.14 & 0.82 & 0.32 \\
Omnidrive~\cite{omnidrive} & VLM & QA & 0.14 & 0.29 & 0.55 & 0.33 & 0.00 & 0.13 & 0.78 & 0.30 \\
OpenDriveVLA\cite{opendrivevla} & VLA &Map\ \&\ Box\ \&\ Motion\ \&\ QA &0.15 &0.31 &0.55 &0.33 &\textbf{0.00}& 0.22& 0.55 &0.25 \\
OccVLA~\cite{occvla} & VLA & QA\ \& \ 3D-Occ\  & 0.18 & 0.26 & 0.40 & 0.28 & - & - & - & - \\

\midrule
\rowcolor{mygray} SparseOccVLA & Unified & QA\ \&\ 3D-Occ & \textbf{0.14} & \textbf{0.22} & \textbf{0.32} & \textbf{0.23} & 0.03 & \textbf{0.12} & \textbf{0.41} & \textbf{0.19} \\
\bottomrule
\end{tabular}
} 
\end{table*}

\subsection{Implementation Details}
\paragraph{Model Details}
We adopt ResNet-50\cite{resnet} to extract multi-scale features from multi-view images of size 256×704, and CLIP-336\cite{clip} vision encoder for distillation. For the occupancy encoder, we randomly initialize 600 queries and 6 encoding layers, where each query generates (2, 4, 16, 32, 56, 112) points across layers. The feature dimension of occupancy query is 1024. We use Vicuna-7B\cite{vicuna} as our LLM, which consists of 32 attention layers with a token feature dimension of 4096. The number of global queries is set to 12. For planning, we cluster 18 anchor trajectories and perform 2 denoising steps for refinement.

\paragraph{Training Details}
Our training process consists of three stages.
In stage 1, we train the sparse occupancy encoder and the connector with occupancy loss and the distillation loss.
In stage 2, we freeze the visual encoder and train the connector, global queries, and downstream forecasting and planning modules.
In stage 3, we train all parameters except for the LLM, which is fine-tuned using the LoRA\cite{lora},  and the LoRA rank $r$ and scaling factor $\alpha$ are set to 128 and 16, respectively.
All training stages are conducted on 8 NVIDIA H20 GPUs with a total batch size of 16. We set the base learning rate to 2e-4 and adopt the AdamW optimizer with a cosine decay schedule to facilitate convergence. The three stages are trained for 36, 6, and 12 epochs, respectively. Each stage directly loads the weights from the previous one.
More details about the model architecture and training settings can be found in our appendix.


\subsection{Main Results}
\paragraph{Scene Understanding}
As shown in the left part of Tab. \ref{tab:understanding and forecasting}, SparseOccVLA significantly outperforms existing state-of-the-art models on Omnidrive-nuScenes\cite{omnidrive} across all three metrics, with a notable 7\% gain CIDEr compared to HERMES\cite{hermes}. Notably, methods based on BEV or image tokens require a large number of tokens, whereas our lightweight version achieves strong performance with only 300 occupancy tokens, demonstrating the high information density. In contrast, OmniDrive-base\cite{omnidrive}, which leverages a Q-Former to compress tokens, suffers from indiscriminate global querying that discards much fine-grained details, resulting in suboptimal performance.
\paragraph{Occupancy Forecasting}
As shown in the right part of Tab. \ref{tab:understanding and forecasting}, SparseOccVLA consistently achieves SOTA performance. With the assistance of the large language model, its occupancy forecasting is markedly improved compared to the similarly sparse SparseWorld\cite{sparseworld}, with an average future 3s mIoU increase of 0.51.

\paragraph{Trajectory Planning}
As presented in Tab. \ref{tab:planning}, we compare SparseOccVLA against end-to-end methods, VLA, and world models. The results demonstrate that SparseOccVLA consistently outperforms all competitors. We argue that with LLM guidance, decoupling anchor scoring from noisy regression allows the model to fully exploit the strengths of both LLMs and diffusion.  The strong performance in planning further demonstrates the representational power of sparse occupancy queries, which can serve as the sole perception-level input.





\subsection{Ablation Studies}
In this section, we provide detailed ablation studies to investigate the contribution of each component and design. Limited by space, we provide additional experimental results in the appendix, including results on nuScenes-QA\cite{nuscenes-qa}, normalization layers for distillation as well as the effects of the number of global queries and image sweeps.

\paragraph{Effect of the Occupancy}

We first investigate the impact of occupancy modeling on overall understanding ability. As shown in 2th row of Tab. \ref{tab:ablation}, when the occupancy supervision is removed and only the textual loss is preserved, the occupancy encoder essentially degenerates into a Q-Former-like module, except that the queries attend to local rather than global features. We observe a significant drop (-0.8) in CIDEr, indicating that the occupancy loss serves as an effective regularizer, endowing the queries with explicit geometric and semantic priors for better LLM understanding. Without it, the heterogeneous and sparse supervision from the LLM alone struggles to achieve effective alignment between the visual encoder and the language model. Furthermore, as shown in the 3th row of Tab. \ref{tab:ablation}, when the 3D position encoding is removed during the alignment from occupancy queries to occupancy tokens, the model fails to converge. This is because, without explicit positional information, the LLM perceives the queries as an unordered and unstructured set of tokens, making it impossible to capture the underlying spatial topology.

\paragraph{Analysis of global queries}
As shown in row 4 of Tab. \ref{tab:ablation}, introducing global queries yields a 0.34 improvement in the CIDEr score, confirming our hypothesis that occupancy queries primarily focus on local details and lack scene-level understanding. The global queries effectively complement this limitation, providing a more holistic representation.

\begin{table}[t]
\centering
\caption{Ablation studies about different components on the understanding metrics and the average mIoU over the future 3s.}
\small
\setlength{\tabcolsep}{4pt}
\renewcommand{\arraystretch}{1.1}
\begin{tabular}{lcccc}
\toprule
\textbf{Method} & CIDEr & METEOR & ROUGE & mIoU \\
\midrule

w/o Occ Encoder        & 0.712 & 0.363 & 0.317 &  -  \\
w/o Position Encoding       & -  & -  & -  &  -  \\
w/o Global Queries  & 0.758 & 0.378 & 0.327 & 13.50 \\

w/o Residual-Query & 0.764 & 0.377 & 0.329 & 13.12 \\
w/o Residual-Token & 0.788 & 0.382 & 0.333 & 13.21 \\
\rowcolor{mygray} Base model & 0.792 & 0.384 & 0.334 & 13.65 \\
\bottomrule
\end{tabular}
\label{tab:ablation}
\end{table}

\begin{figure}
    \centering
    \includegraphics[width=1.0\linewidth]{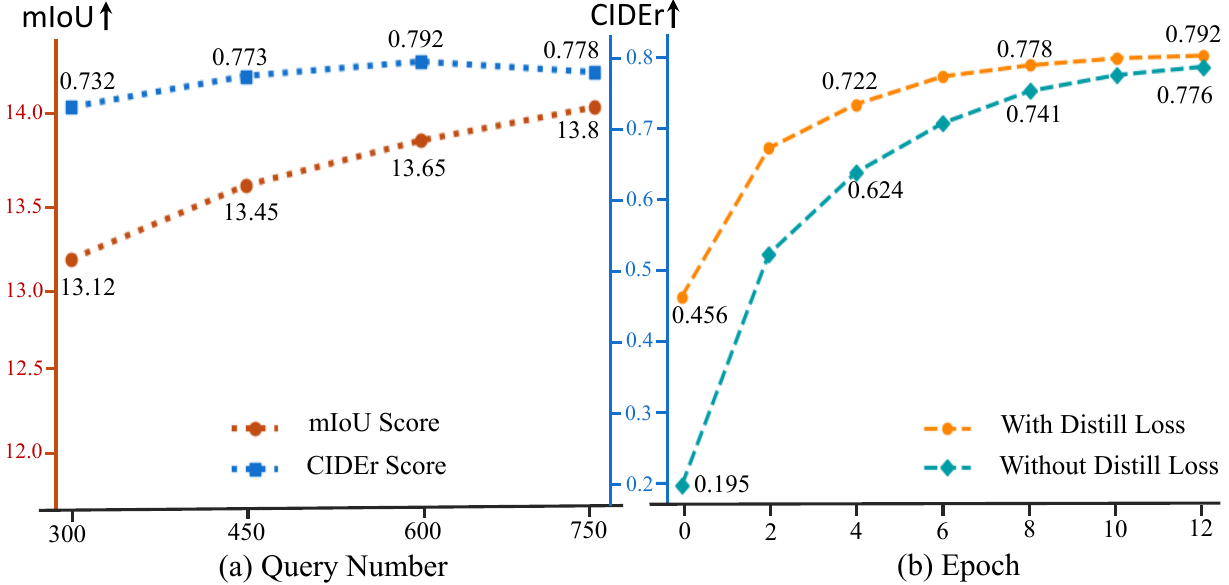}
    \caption{Ablation study on the number of queries (a), and the distillation loss (b).}
    \label{fig:ablation}
\end{figure}

\paragraph{Residual Fusion and Uniformity Analysis}
As shown in row 5 of Tab. \ref{tab:ablation}, removing the residual fusion of occupancy query causes a noticeable drop in CIDEr (-0.28). We hypothesize that this is because the queries must simultaneously align with the LLM while preserving their original perception-level features, thus increasing the LLM’s learning burden. In row 6, when the occupancy tokens are removed, the LLM becomes decoupled from forecasting and is only responsible for text-level understanding. In this case, the language metrics remain similar to the standard model, but the mIoU remains low (-0.44). This indicates that LLM understanding facilitates occupancy forecasting, while the reverse effect is weaker. By effectively fusing high-level occupancy tokens with low-level occupancy queries, SparseOccVLA leverages the complementary strengths of both, highlighting the advantage of unified modeling.




\paragraph{Effect of Query Number}
As shown in the {\color[HTML]{1370CC} blue} curve on the Fig. \ref{fig:ablation}(a), SparseOccVLA exhibits insensitivity to the number of queries. Even with only 300 queries, it still achieves a CIDEr score of 0.732, demonstrating its high information density and robust scene understanding capability. We also observe a decline in CIDEr when the number of queries increases, which we attribute to the increased disorder that imposes a heavier interpretative burden on the LLM. In contrast, as indicated by the {\color[HTML]{C14D12} red} curve, the mIoU shows a consistent positive correlation with the number of queries due to its low-level nature.

\paragraph{Effect of the Distillation Loss}
As shown in the Fig. \ref{fig:ablation} (b), applying the distillation loss during the training of the Occupancy Encoder immediately yields an initial CIDEr score of 0.456 and consistently outperforms the counterpart without distillation. Although the final performance of both models eventually converges, introducing the distillation loss significantly accelerates convergence and enhances training stability. We observe that removing the distillation loss often causes gradient explosion and instability, particularly with a large number of queries, while its inclusion effectively mitigates such failures.
\begin{figure*}
    \centering
    \includegraphics[width=1\linewidth]{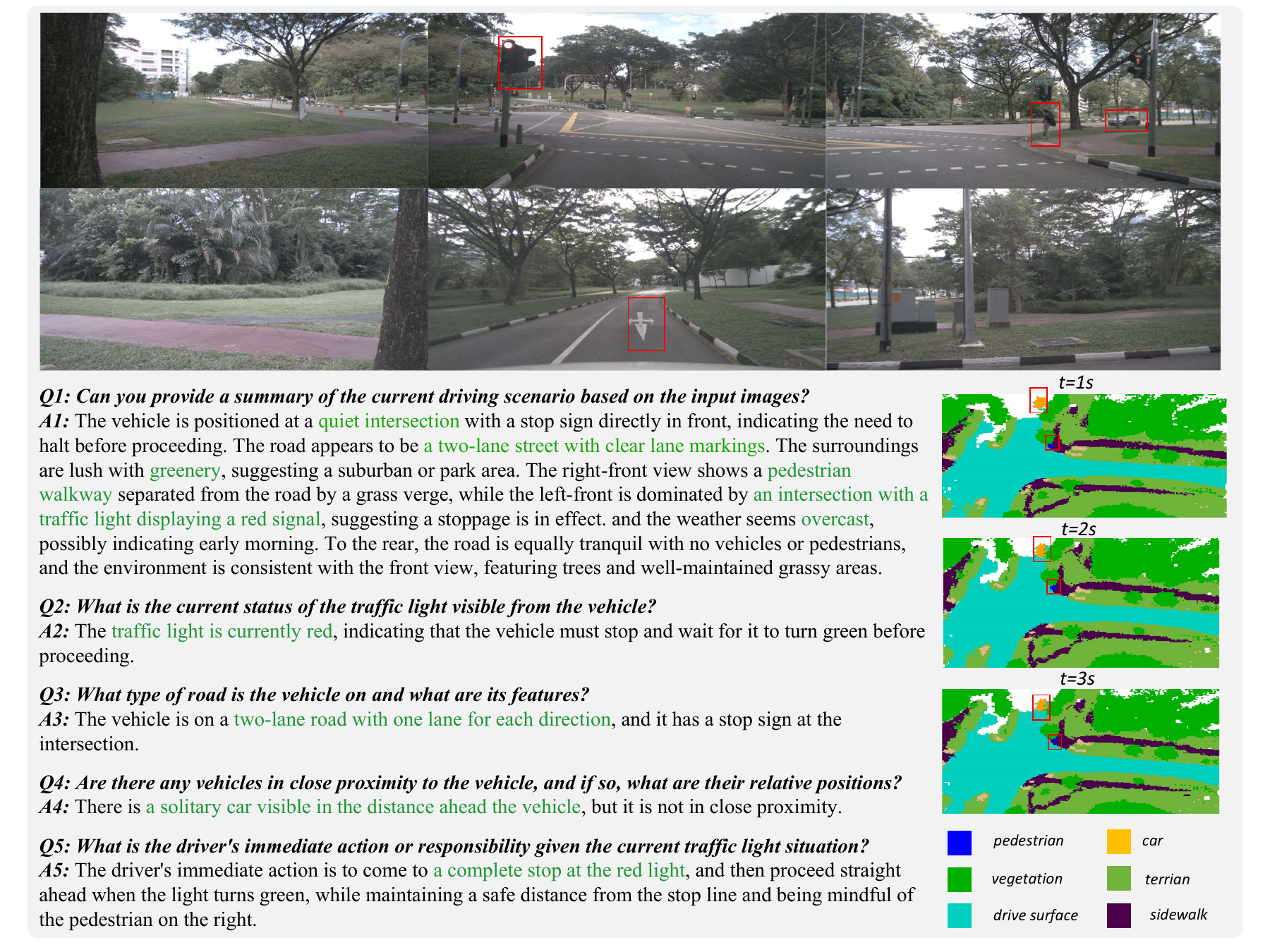}
    \caption{Qualitative results of SparseOccVLA for scene understanding and occupancy forecasting.}
    \label{fig:visualization}
\end{figure*}

\paragraph{Effect of LLM-guided Planning}
\begin{table}[t]
\centering
\caption{Ablation study on the planning module. 
We report L2 distance error and collision rate at 1s, 2s, 3s, and their averages.}
\small
\setlength{\tabcolsep}{4pt}
\begin{tabular}{lcccccc}
\toprule
\multirow{2}{*}{\textbf{Model Setting}} & 
\multicolumn{3}{c}{\textbf{L2 (m)}} & 
\multicolumn{3}{c}{\textbf{Coll. Rate (\%)}} \\
\cmidrule(lr){2-4} \cmidrule(lr){5-7}
& 1s & 2s & 3s & 1s & 2s & 3s  \\
\midrule
w/o LLM Scoring & 0.17 & 0.28 & 0.41  & 0.03 & 0.25 & 0.49  \\

w/o Traj-Reason Fusion & 0.14 & 0.26 & 0.35  & 0.04 & 0.19 & 0.41\\
w/o Traj-Scene Fusion & 0.15 & 0.28 & 0.38  & 0.04 & 0.23 & 0.46  \\
w/o Traj-Ego Fusion & 0.22 & 0.35 & 0.68  & 0.07 & 0.29 & 0.65  \\
\rowcolor{mygray} Base Model & 0.14 & 0.22 & 0.30  & 0.03 & 0.16 & 0.39  \\
\bottomrule
\end{tabular}
\label{tab:planning_ablation}
\end{table}
As shown in Tab. \ref{tab:planning_ablation}, row 3, removing the LLM-guided scoring—leaving the planner to independently score while regressing anchor noise, results in a notable degradation in both L2 distance and collision rate metrics. We attribute this to SparseOccVLA effectively degenerating into a purely end-to-end planner, lacking a global perspective and imposing a greater optimization burden on the model. LLMs, in contrast, offer robust high-level scene understanding and reasoning capacity.

\paragraph{Effect of Multi-Modal Fusion}
As shown in rows 4–6 of Tab. \ref{tab:ablation}, fusing the trajectory embeddings with language instructions, scene context, and ego-state reduces the average L2 error by 0.03m, 0.05m, and 0.19m, respectively. This demonstrates the consistent benefits of cross-modal information interaction for accurate trajectory regression. Specifically, reasoning tokens provide subjective decision context, occupancy queries imposes scene-level constraints, and ego-state further refines trajectory quality.

\subsection{Qualitative Results}
We qualitatively visualize representative results of scene understanding and forecasting in Fig. \ref{fig:visualization}. It is evident that SparseOccVLA successfully perceives pedestrians and distant vehicle, which are faithfully reflected in the LLM outputs. It also accurately recognizes red traffic lights and lane markings, benefiting from its fully end-to-end design that enables effective capture of non-geometric visual cues.
\section{Conclusion}
In this work, we propose SparseOccVLA, which, to the best of our knowledge, is the first to effectively integrate VLMs and occupancy representations, achieving strong performance across understanding, occupancy forecasting, and planning. With only sparse occupancy queries serving as the sole visual bridge to the LLM—without relying on visual tokens or other perceptual contents—SparseOccVLA establishes a new paradigm beyond MLP-, Q-Former-, and BEV-based alignment strategies that will foster cross-domain research and advances the community.
\paragraph{Limitation}
Despite its strong performance, SparseOccVLA has several limitations. It relies on dense occupancy supervision, leading to non-trivial construction cost. Moreover, since the closed-loop benchmarks lacks occupancy ground truth, its closed-loop planning capability remains to be further evaluated in future work.


{
\small
    \bibliographystyle{ieeenat_fullname}
    \bibliography{main}
}
\appendix
\setcounter{page}{1}
\maketitlesupplementary

\section{More Details}
\paragraph{Model Details}
To endow SparseOccVLA with dynamic forecasting capability, our Sparse Query Encoder incorporates the Adaptive Scaling in \cite{sparseworld} that adjusts the initial query range based on ego states. Each query learns 4 sampling points, which are projected onto and interpolated from the 4-scale feature maps extracted by the visual encoder, followed by the adaptive mixing\cite{sparsebev} to consolidate the information. To enhance representational capacity, the occupancy queries are configured with a feature dimension of 1024 and 32 attention heads.

In contrast to standard Multi-Head Self-Attention, the spatially grounded MHSA used in our encoder computes attention weights as follows:
\begin{equation}
    A_{ij} = q_i^Tq_j-\tau_i||\mathbf{p}_i-\mathbf{p}_j||,
\end{equation}
where $i,j=1,2,...,N$, $p_i$ index the queries, and $\tau_i$ is learned from occupancy query $q_i$.

Following OmniDrive\cite{omnidrive}, we also encode the ego state into a single token and feed it to the language model for interpretation. In our anchor-diffusion planner, the hidden feature dimension is aligned with that of the occupancy queries, i.e., $D=1024$.
\paragraph{Sparse Occupancy Encoder Training}
To adapt to the occupancy forecasting task, the Adaptive Scaling strategy in \cite{sparseworld} is adopted during the training of the Sparse Occupancy Encoder to enlarge the initial query range for dynamic environments. We concatenate the current and future $T$ frames of occupancy ground truth and perform coordinate transformations according to the ego pose to obtain extended-range occupancy supervision. 

At each encoding layer, Chamfer Distance loss is employed for geometric supervision, while Focal loss provides semantic supervision, jointly guiding the model learning process. The overall training objective of the Sparse Occupancy Encoder is defined as follows:
\begin{equation}
\mathcal{L}_{\text{occ}} 
= \sum_{l=0}^{L} 
\, \mathcal{L}_{\text{CD}}(\mathcal{P}_l, \mathcal{G})
+ 
\sum_{l=1}^{L}\lambda_{\text{focal}} \, \mathcal{L}_{\text{focal}}(C_l, \hat{C}_l) + \mathcal{L}_{distill},
\end{equation}
where $\mathcal{L}_\text{CD}(\mathcal{P}_0, \mathcal{G})$ denotes that the geometric supervision is also applied to the original point set $\mathcal{P}_0$. $C_l$ and $\hat{C}_l$ represent the predicted point classes and their corresponding ground truth labels at layer $l$, respectively. The weighting factor $\lambda_{focal}$ is set to 0.2.

\paragraph{End to End Training}
During the end-to-end training phase, all modules are jointly optimized. The occupancy forecasting is supervised on a per-frame basis. Unlike \cite{sparseworld}, we utilize the concatenated occupancy ground truth $\mathcal{G}$ as the base and transform the supervision to the corresponding future frames according to expert trajectories using the vehicle pose matrices. The overall occupancy prediction loss is computed as follows:
\begin{equation}
\mathcal{L}_{\text{fore}} 
= \sum_{t=1}^{T} 
\, \mathcal{L}_{\text{CD}}(\mathcal{P}_t, \mathcal{G}_t)
+ 
\sum_{t=1}^{T} \lambda_{\text{focal}} \, \mathcal{L}_{\text{focal}}(C_t, \hat{C}_t),
\end{equation}
where $T$ denotes the number of predicted frames, and $\mathcal{P}_t, \mathcal{G}_t, C_t, \hat{C}_t$ represent the predicted point set, the transformed ground truth, the predicted point classes, and the corresponding ground truth labels at time step $t$, respectively.

The total training objective consists of four components, namely the occupancy perception loss, the language modeling loss, the occupancy prediction loss, and the trajectory noise prediction loss:

\begin{equation}
    \mathcal{L}_{\text{total}} = \mathcal{L}_{\text{occ}}+\mathcal{L}_{\text{LM}}+\mathcal{L}_{\text{for}e}+\mathcal{L}_{\text{diff}}
\end{equation}

\section{More Analysis}
\begin{figure*}
    \centering
    \includegraphics[width=1\linewidth]{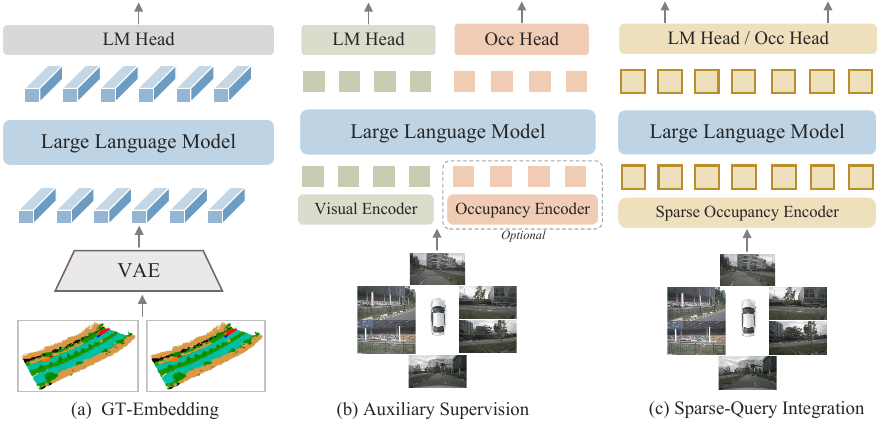}
    \caption{Comparison of different occupancy-VLM integration paradigms.}
    \label{fig:comparison}
    \label{fig:placeholder}
\end{figure*}
In this section, we analyze the significance of SparseOccVLA to the autonomous driving community and explain why we regard it as the first effective integration of vision-language models and semantic occupancy.

As discussed in the main text, both paradigms have achieved remarkable progress in recent years—VLMs excel in high-level reasoning, while occupancy representations provide precise, fine-grained spatial understanding. Combining these two paradigms is naturally appealing, as they possess complementary strengths. However, this integration is highly challenging due to the severe cross-modal gap and computational inefficiency. Occupancy features are dense and low-level, while language tokens are sparse and abstract, making alignment inherently difficult. Regarding the role of occupancy in Vision-Language Models (VLMs), we categorize existing methods into three types, as illustrated in Fig. \ref{fig:comparison}.

As illustrated in Fig. \ref{fig:comparison}(a), early explorations\cite{occllm,occllama} attempted to embed occupancy ground truth via a VAE before feeding it into a large language model. Although alleviating computational cost, it completely decouples upstream perception from downstream reasoning. Consequently, the LLM only observes geometrically defined elements while discarding critical non-geometric cues such as traffic lights, lane markings, and signs, which limit the model’s upper bound. This design violates the principle of end-to-end autonomous driving and has not been widely recognized.

More recent works\cite{occvla}, shown in Fig. \ref{fig:comparison}(b), introduce an auxiliary occupancy encoder alongside the vision encoder for additional supervision. However, such designs still depend on visual tokens and thus suffer from the same limitations as conventional VLMs. In these methods, occupancy plays only a marginal role rather than fundamentally departing from the established VLM paradigm.

In contrast, SparseOccVLA (Fig. \ref{fig:comparison}(c)) introduces a sparse occupancy encoder that extracts sparse occupancy queries serving as the sole bridge between vision and language, completely eliminating the dependency on any visual tokens.

These sparse occupancy queries offer three key advantages:
\begin{itemize}
    \item \textbf{Compact}: leveraging the efficiency of sparse-based occupancy prediction. In our implementation, only 300 tokens are required to achieve outstanding scene understanding performance.
    \item \textbf{Information rich}: preserving dense geometric and semantic cues from raw visual inputs, aligning with the principle of end-to-end modeling.
    \item \textbf{Modeling friendly}: possessing explicit spatial attributes that enable flexible temporal fusion, enhancing spatial reasoning and interpretability.
\end{itemize}

In summary, SparseOccVLA effectively resolves long-standing issues of information loss, external dependency, and computational inefficiency, achieving a truly unified framework that combines the complementary advantages of occupancy and VLMs in a principled and efficient manner.

\section{More Experiments}
\subsection{Results on nuScenes-QA}
Following previous works\cite{omnidrive}, we conduct experiments on the nuScenes-QA\cite{nuscenes-qa} dataset to demonstrate the generalization capability of SparseOccVLA. nuScenes-QA is another widely used visual question answering (VQA) benchmark. Unlike OmniDrive-nuScenes\cite{omnidrive}, which mainly contains long-form textual answers, the answers in nuScenes-QA are mostly short and deterministic words. As shown in Tab. \ref{tab:nuscenes-qa}, SparseOccVLA also achieves an accuracy of 60.8\%, significantly surpassing the existing state-of-the-art methods.
\begin{table}[t]
  \centering
  \small
  \caption{Comparison of different methods on nuScenes-QA benchmark.}
  \label{tab:nuscenes-qa}
  \begin{tabular}{lccc}
    \toprule
    \textbf{Method} & \textbf{Reference} & \textbf{Input} & \textbf{Acc.(\%)}$\uparrow$ \\
    \midrule
    LLaVA\cite{llava} & NeurIPS 23 & Image & 47.4 \\
    LiDAR-LLM\cite{lidar-llm} & AAAI 25 & LiDAR & 48.6 \\
    BEVDet+MCAN\cite{nuscenes-qa} & AAAI 24 & Image & 57.9 \\
    CenterPoint+MCAN\cite{nuscenes-qa} & AAAI 24 & LiDAR & 59.5 \\
    OmniDrive\cite{omnidrive} & CVPR 25 & Image & 59.2 \\
\rowcolor{mygray}    SparseOccVLA & Ours & Image & 60.8 \\
    \bottomrule
  \end{tabular}
  
\end{table}

\subsection{More Ablation Results}
In this section, we provide additional ablation studies to illustrate the detailed design of our model and help readers gain a deeper understanding of our approach.

\paragraph{Effect of Normalization Layers}

\begin{table}[ht]
  \centering
  \small
  \renewcommand{\arraystretch}{1.2}
    \caption{Ablation study on the effect of normalization layer. We report the mIoU perceptual score of the current frame, which reflect the learning effectiveness of the Sparse Occupancy Encoder.}
  \label{tab:norm_layer}
  \begin{tabular}{lccccc}
    \toprule
    \textbf{Distill} & \textbf{Norm} & \textbf{mIoU}& \textbf{CIDEr} & \textbf{METEOR} & \textbf{ROUGE} \\
    \midrule
    \ding{55} & \ding{55} & 17.68  &  0.776  &  0.376  &  0.329  \\
    \ding{51} & \ding{55} & 15.43  &  0.743  &  0.372  &  0.324  \\
\rowcolor{mygray}    \ding{51} & \ding{51} & 17.65  &  0.792  & 0.384   &  0.334  \\
    \bottomrule
  \end{tabular}

\end{table}

As shown in Tab. \ref{tab:norm_layer}, we present the ablation study on the effect of the normalization layer. When the Norm Layer is removed, the mIoU of the current frame drops significantly compared to the setting with the Norm Layer, which further degrades the language understanding metrics. We attribute this to the overly strong distillation loss that enforces the occupancy features to be directly aligned with the full language space, leading to gradient conflicts and hindering the learning of the Sparse Occupancy Encoder. In contrast, the Norm Layer effectively alleviates this issue by enabling smoother feature alignment without imposing excessive fine-grained constraints. Comparing the 2nd and 4th rows, together with other ablations in the main text, we observe that the Norm Layer greatly improves training stability and convergence speed while maintaining the perception capability of the occupancy encoder.

\paragraph{Effect of the Number of Global Queries}

\begin{table}[ht]
\centering
\small
\caption{Effect of different numbers of global queries on various language metrics.}
\label{tab:global_query}
\begin{tabular}{cccc}
\toprule
\textbf{Global Query} & \textbf{CIDEr} & \textbf{METEOR} &\textbf{ROUGE} \\
\midrule
0   & 0.758 & 0.378 & 0.327  \\
4  & 0.775 & 0.381 & 0.330  \\
8   & 0.786 & 0.383 & 0.332 \\
\rowcolor{mygray}12  & 0.792 & 0.384 & 0.333 \\
16  & 0.793 & 0.384 & 0.334 \\
\bottomrule
\end{tabular}

\end{table}
Tab. \ref{tab:global_query} presents an analysis of SparseOccVLA's sensitivity to the number of global queries. Overall, increasing the number of global queries significantly enhances the model’s understanding capabilities, exhibiting a clear scaling trend. Performance saturates when the number of global queries exceeds 12, but no degradation is observed. This observation supports our hypothesis: while sparse occupancy queries focus primarily on local details, global queries attend to and integrate information from all queries, thereby facilitating the LLM’s comprehension of the entire scene.

\paragraph{Effect of the Image Frames}
\begin{table}[htbp]
\centering
\small
\caption{Impact of the number of input image frames on understanding and prediction performance of SparseOccVLA.}
\begin{tabular}{cccccc}
\toprule
\multirow{2}{*}{Frames} & \multicolumn{3}{c}{Understanding} & \multicolumn{2}{c}{mIoU} \\
\cmidrule(lr){2-4}\cmidrule(lr){5-6}
 & CIDEr & METEOR & ROUGE & Curr & Fut.avg \\
\midrule
1  & 0.742 & 0.381 & 0.326 & 16.52 & 12.60 \\
2  & 0.759 & 0.385 & 0.30 & 16.97 & 13.02 \\
4  & 0.776 & 0.387 & 0.331 & 17.38 & 13.31 \\
\rowcolor{mygray}8  & 0.792 & 0.388 & 0.334 & 17.65 & 13.53 \\
\bottomrule
\end{tabular}
\label{tab:frame_sensitivity}
\end{table}

SparseOccVLA inherits the advantages of the Sparse Occupancy Encoder, enabling it to attend to historical image frames in parallel and thereby exhibiting flexible and powerful temporal modeling capabilities. In Table 4, we investigate the impact of the number of input image frames on both understanding and prediction performance. We observe a positive correlation between the number of input frames and the performance in scene understanding as well as occupancy perception and forecasting, reflecting the model’s scalability. Notably, lower-level mIoU metrics demonstrate higher sensitivity to the number of frames. Traditional vision-language models struggle to process video streams, which limits the language model’s ability to capture temporal dynamics and constrains the overall performance ceiling. We also observe that even with a single input frame, although performance is significantly reduced, SparseOccVLA still performs comparably to the baseline HERMES\cite{hermes} model, demonstrating the potential of sparse occupancy queries to replace BEV representations.

\paragraph{Effect of Occupancy Query Dimension}
\begin{table}[t]
\centering
\small
\caption{Effect of the occupancy query feature dimension $D$ on model performance.}
\label{tab:demension}
\begin{tabular}{cccc}
\toprule
\textbf{Feature Dim ($D$)} & \textbf{CIDEr} & \textbf{METEOR} & \textbf{ROUGH} \\
\midrule
256  & 0.761 & 0.375 & 0.327 \\
512  & 0.779 & 0.383 & 0.332 \\
\rowcolor{mygray}1024 & 0.792 & 0.388 & 0.334 \\
\bottomrule
\end{tabular}
\end{table}
In Tab. \ref{tab:demension}, we investigate the influence of the occupancy-query feature dimension $D$. Notably, $D$ is decoupled from both the image encoder and the LLM, allowing it to be flexibly adjusted. Unlike precedent works\cite{opus,sparseworld} that typically fixes $D=256$, our early experiments indicate that such a low-dimensional setting leads to suboptimal semantic understanding. As shown in Tab. \ref{tab:demension}, increasing the occupancy-query dimension from 256 to 1024—while keeping all other hyperparameters unchanged—consistently improves performance. This trend demonstrates the scalability of our model.

\paragraph{Experiments of other LLMs}
To ensure comprehensive comparison with existing methods and to assess the generalization of SparseOccVLA, we replace Vicuna\cite{vicuna} with InternVL2\cite{internvl} and retrain as well as reevaluate the model. In our implementation, the visual encoder of InternVL is frozen and substituted with our sparse occupancy query. As shown in the Tab. \ref{tab:model size}, under the same parameter budget, SparseOccVLA consistently outperforms HERMES\cite{hermes} across language metrics, further demonstrating the effectiveness of the proposed sparse occupancy query design.

\begin{table}[t]
\centering
\caption{Comparison of language metrics across different model sizes.}
\small
\setlength{\tabcolsep}{5pt}  
\label{tab:model size}
\begin{tabular}{lcccc}
\toprule
\textbf{Model} & \textbf{LLM} & \textbf{CIDEr} & \textbf{METEOR} & \textbf{ROUGE} \\
\midrule
HERMES       & InternVL2-2B & 0.377 & 0.321 & 0.740 \\
\rowcolor{mygray} Ours & InternVL2-2B & 0.386 & 0.335 & 0.785 \\
\midrule
HERMES       & InternVL2-4B & 0.381 & 0.325 & 0.747 \\
\rowcolor{mygray} Ours & InternVL2-4B & 0.388 & 0.338 &0.797  \\
\bottomrule
\end{tabular}
\end{table}

\paragraph{Comparison of Anchor Number and Denoising Steps}
\begin{table}[t]
\centering
\caption{Effect of denoising steps and anchor numbers on planner performance.}
\small
\label{tab:planner_ablation_fixed}
\begin{tabular}{cccccc}
\toprule
\multirow{2}{*}{\textbf{Denoising}} & \multirow{2}{*}{\textbf{Anchor }} & \multicolumn{4}{c}{\textbf{L2 Error (m)}} \\
\cmidrule(lr){3-6}
 & & 1s & 2s & 3s & Avg \\
\midrule
1 & 18 & 0.145 & 0.237 & 0.332 & 0.238 \\
\rowcolor{mygray}2 & 18 & 0.141 & 0.224 & 0.298 & 0.221 \\
3 & 18 & 0.140 & 0.224 & 0.300 & 0.221 \\
2 & 15 & 0.146 & 0.247 & 0.348 & 0.247 \\
2 & 21 & 0.142 & 0.229 & 0.313 & 0.228 \\
\bottomrule
\end{tabular}
\end{table}

In Tab. \ref{tab:planner_ablation_fixed}, we analyze the sensitivity of SparseOccVLA’s planning metrics with respect to the number of denoising steps and the number of anchors. The results show that the model is relatively robust to the number of denoising steps: only two denoising steps are sufficient to achieve the best performance, and additional steps do not yield further improvements. In contrast, SparseOccVLA is more sensitive to the number of anchors. Following the design of DiffusionDrive\cite{diffusiondrive}, we observe that generating six anchors per direction—resulting in a total of 18 anchors—produces the highest planning scores. We attribute this trend to the fact that too few anchors fail to adequately cover the entire scene, while too many anchors introduce mutual interference.

\section{Efficiency Comparison}
In this section, we provide a detailed analysis of the training and inference efficiency of SparseOccVLA, and compare it against state-of-the-art methods.
\paragraph{Model Analysis}
As shown in the Tab. \ref{tab:model_comparison}, we compare SparseOccVLA with other models in terms of parameters outside the language model. Our visual backbone uses the lightweight ResNet-50\cite{resnet} to preserve perception-level details. Moreover, the input image resolution is significantly lower than that of OmniDrive and HERMES. Such a lightweight perception module directly contributes to improved training and inference efficiency. We provide a detailed analysis in the following.

\begin{table}[t]
\centering
\caption{Detailed Comparison of model configurations with vision components.}
\small
\label{tab:model_comparison}
\begin{tabular}{lccccc}
\toprule
\textbf{Model} & \textbf{Vision} & \textbf{Token}  & \textbf{Image-Size} \\
\midrule
OmniDrive\cite{omnidrive}      & EVA-02-L\cite{eva-v2} & 513 &    $640\times640$ \\
HERMES\cite{hermes}         & OpenCLIP\cite{clip} & 2500 &   $1600\times900$  \\
\rowcolor{mygray} SparseOccVLA & ResNet-50 & 612 &$256\times704$  \\
\bottomrule
\end{tabular}
\end{table}

\paragraph{Training Efficiency}
\begin{table}[t]
\centering
\caption{Comparison of training efficiency and hardware usage.}
\small
\label{tab:training_efficiency}
\begin{tabular}{lcccc}
\toprule
\textbf{Model} & \textbf{Device}& \textbf{Epoch} & \textbf{Time} & \textbf{Memory} \\
\midrule
HERMES\cite{hermes}        &H20\ $\times\ 32$ & 36 & 26h & 80G \\
\rowcolor{mygray} SparseOccVLA &H20\ $\times\ 8$ & 12 & 11h & 45G \\
\bottomrule
\end{tabular}
\end{table}
As shown in Tab. \ref{tab:training_efficiency}, we compare the training resources required by HERMES\cite{hermes} and our SparseOccVLA, where both methods utilize InternVL2-2B\cite{internvl} as the language model and adopt the official default settings for all other components. We observe that SparseOccVLA is highly efficient: it converges within only 12 epochs using 8 H20 GPUs, with a total training time less than half of HERMES and a peak memory usage slightly above half of it. This efficiency primarily stems from SparseOccVLA’s lightweight visual encoder and its highly compact token representation.

\section{More Qualitative results}
\begin{figure*}
    \centering
    \includegraphics[width=1\linewidth]{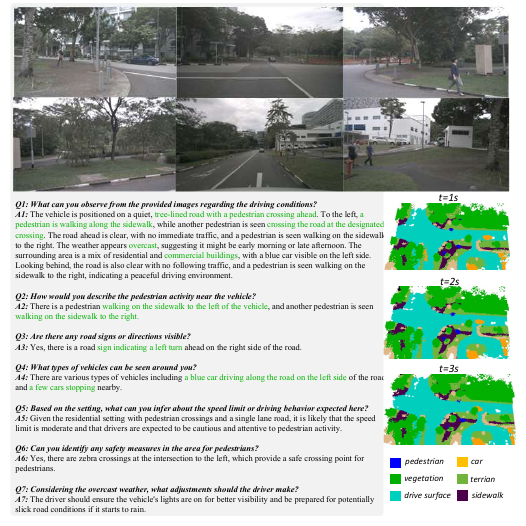}
    \caption{A visualization of reasoning results in an intersection scenario.}
    \label{fig:vis1}
\end{figure*}

\begin{figure*}
    \centering
    \includegraphics[width=1\linewidth]{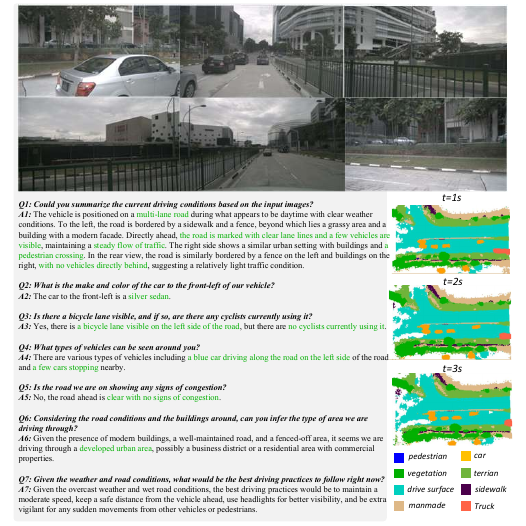}
    \caption{A visualization of reasoning results in an urban road scenario.}
    \label{fig:vis2}
\end{figure*}

\begin{figure*}
    \centering
    \includegraphics[width=1\linewidth]{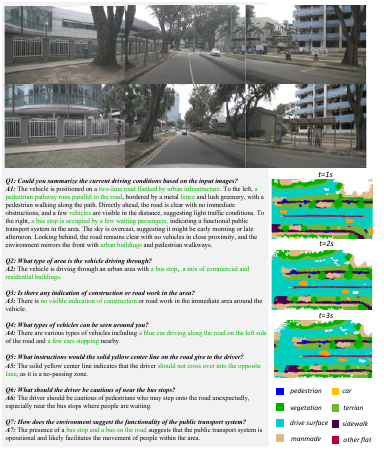}
    \caption{A visualization of reasoning results in an urban road scenario.}
    \label{fig:vis3}
\end{figure*}

In Fig. \ref{fig:vis1}, Fig. \ref{fig:vis2}, and Fig. \ref{fig:vis3}, we present additional visualization results of SparseOccVLA’s understanding and forecasting capabilities. The analysis reveals that SparseOccVLA not only reconstructs driving scenes effectively and comprehends the environment in textual form, but also performs deep reasoning and analysis, demonstrating its powerful multimodal abilities. These visualizations also highlight its ability to anticipate future interactions.

\end{document}